\documentclass[sigconf]{aamas} 

\usepackage{balance} 

\setcopyright{ifaamas}
\acmConference[AAMAS '23]{Proc.\@ of the 22nd International Conference
	on Autonomous Agents and Multiagent Systems (AAMAS 2023)}{May 29 -- June 2, 2023}
{London, United Kingdom}{A.~Ricci, W.~Yeoh, N.~Agmon, B.~An (eds.)}
\copyrightyear{2023}
\acmYear{2023} 
\acmDOI{}
\acmPrice{}
\acmISBN{}

\acmSubmissionID{28}

\title[Minimax Strikes Back]{Minimax Strikes Back}

\author{Quentin Cohen-Solal}
\affiliation{
	\institution{LAMSADE, Université Paris-Dauphine, PSL, CNRS}
	\city{Paris}
	\country{France}}
\email{quentin.cohen-solal@dauphine.psl.eu}

\author{Tristan	Cazenave}
\affiliation{
	\institution{LAMSADE, Université Paris-Dauphine, PSL, CNRS}
	\city{Paris}
	\country{France}}
\email{tristan.cazenave@dauphine.psl.eu}

\usepackage{multirow}

\usepackage{comment}

\usepackage{url}
\usepackage{amsfonts}
\usepackage{nicefrac}
\usepackage{xcolor}

\usepackage[T1]{fontenc}
\usepackage[utf8]{inputenc}
\usepackage{units}
\usepackage{multirow}
\usepackage{algorithm2e}
\usepackage{graphicx}
\usepackage{microtype}
\usepackage{soul}
\usepackage{url}
\usepackage{caption}

\ccsdesc[500]{Computing methodologies~Reinforcement learning}
\ccsdesc[500]{Computing methodologies~Game tree search}
\ccsdesc[500]{Computing methodologies~Heuristic function construction}
\ccsdesc[100]{Computing methodologies~Planning for deterministic actions}
\ccsdesc[100]{Computing methodologies~Multi-agent planning}
\ccsdesc[100]{Computing methodologies~Neural networks}

\keywords{Minimax ; Reinforcement Learning ; Zero Learning ; Games ; Tree Search}

\newcommand{\BibTeX}{\rm B\kern-.05em{\sc i\kern-.025em b}\kern-.08em\TeX}

\makeatletter

\providecommand{\tabularnewline}{\\}

\theoremstyle{plain}
\newtheorem{thm}{\protect\theoremname}
\theoremstyle{definition}
\newtheorem{defn}[thm]{\protect\definitionname}
\theoremstyle{plain}

\theoremstyle{plain}

\global\long\def\argmax{{\mathrm{arg\,max}}}%
\global\long\def\argmin{{\mathrm{arg\,min}}}%

\global\long\def\id{\mathrm{ID}}%
\global\long\def\ubfm{\mathrm{UBFM}}%
\global\long\def\ubfms{\ubfm_{s}}%
\global\long\def\mcts{\mathrm{MCTS}}%

\makeatother

\providecommand{\definitionname}{Definition}

\begin{abstract}
Deep Reinforcement Learning reaches a superhuman level of play in
many complete information games. The state of the art algorithm for
learning with zero knowledge is AlphaZero. We take another approach,
Athénan, which uses a different, Minimax-based, search
algorithm called Descent, as well as different learning targets and that does not
use a policy. We show that for multiple games it is much more efficient
than the reimplementation of AlphaZero: Polygames. It is even competitive
with Polygames when Polygames uses 100 times more GPU (at least for
some games). One of the keys to the superior performance is that the
cost of generating state data for training is approximately 296 times
lower with Athénan. With the same reasonable ressources,
Athénan without reinforcement heuristic is at least $7$ times
faster than Polygames and much more than $30$ times faster with reinforcement
heuristic.
\end{abstract}

\begin{document}
	
	
	\pagestyle{fancy}
	\fancyhead{}
	
	
	\maketitle

\global\long\def\terminalrandom{\mathrm{t_{r}}}%
\global\long\def\hfb{\mathrm{b_{t}}}%
\global\long\def\hfp{\mathrm{p_{t}}}%
\global\long\def\hadapt{f_{\theta}}%
\global\long\def\hadaptnum#1{f_{\theta_{#1}}}%
\global\long\def\itreeset{\mathbf{T_{i}}}%
\global\long\def\treeset{\mathbf{T}}%
\global\long\def\rootset{\mathbf{R}}%
\global\long\def\hterminal{f_{\mathrm{t}}}%
\global\long\def\rnap{\mathrm{p_{rna}}}%
\global\long\def\rnab{\mathrm{b_{rna}}}%
\global\long\def\ubfm{\mathrm{UBFM}}%
\global\long\def\ubfmt{\ubfm_{\mathrm{s}}}%
\global\long\def\argmax{\operatorname*{\mathrm{arg\,max}}}%
\global\long\def\argmin{\operatorname*{\mathrm{arg\,min}}}%
\global\long\def\liste#1#2{\left\{  #1\,|\,#2\right\}  }%
\global\long\def\fterminal{\hterminal}%
\global\long\def\fadapt{\hadapt}%
\global\long\def\id{\mathrm{ID}}%
\global\long\def\minimum{\operatorname*{\mathrm{min}}}%

\section{Introduction}

Monte Carlo Tree Search (MCTS) \citep{Coulom2006,Kocsis2006,BrownePWLCRTPSC2012}
and its refinements \citep{cazenave2015grave,CazenavePPAF16,silver2016mastering}
are the current state of the art in complete information games search
algorithms. Historically, at the root of MCTS were random and noisy
playouts. Many such playouts were necessary to accurately evaluate
a state. Since \emph{AlphaGo} \citep{silver2016mastering} and \emph{AlphaZero}
\citep{silver2018general} it is not the case anymore. More precisely, strong policies
and evaluations are now provided, by neural networks that are trained
with Reinforcement Learning. These evaluations are stronger and faster to calculate. 

In AlphaGo and its descendants the policy
is used as a prior in the PUCT bandit to explore first the most promising
moves advised by the neural network policy. In addition, the neural evaluations
replace the playouts. Moreover, in AlphaGo, before the reinforcement learning process,  data from matches played between humans are used during a supervised learning process. It is not the case with the latest version, i.e. AlphaZero, where a very
high level of play can be achieved without the use of knowledge. For example, 
AlphaZero has surpassed the level of the program Stockfish 8
in Chess (Grandmaster level) \citep{silver2018general}. In this paper,
we advocate that with reinforcement learning, MCTS might not be the
best algorithm anymore. Minimax algorithms are serious challengers
when equipped with a strong evaluation function from reinforcement
learning.

In this paper, we show that minimax-based algorithms are competitive with MCTS-based algorithms, and even superior for at least many games.
More precisely,  we make a comparison between a recent Minimax-based
reinforcement learning framework, called \emph{Athénan}, with AlphaZero,
the state of the art of reinforcement learning, which had not been
done before. Unlike the AlphaZero approach based on MCTS, Athénan uses two search
algorithms, which are variants of Minimax: \emph{Descent} used during the learning process and \emph{Unbounded Minimax} used after the learning process.

The remainder of the paper is organized as follows. The second section
deals with related work. In particular, the section \ref{learning}
presents the two learning algorithms we use in this paper: Athénan and AlphaZero, but also the open source reimplementation of the latter named \emph{Polygames}. This section also compares their characteristics. Sections
\ref{sec:compar_learning} experimentally
compares the two learning algorithms. Section \ref{sec:discusion}
is a discussion about the results and Section \ref{sec:conclusion}
concludes the article.
Important note: Details of experiments, removed for a reason of space,
are described in Technical Appendix  \cite{cohen2023appendix}, which includes the description
of the used games, details about the used neural architectures, the
learning parameters, algorithms, other experiments, and additional performance curves and tables.

\section{Background}

\subsection{Related Work}

There are many search algorithms for perfect information games. The
two standard algorithms are Monte Carlo Tree Search and Minimax with
$\alpha\beta$  pruning.

MCTS has its roots in computer Go \citep{Coulom2006}. It was theoretically
defined with the UCT algorithm \citep{Kocsis2006} that converges to
the Nash equilibrium and uses a well defined bandit, Upper Confidence
Bounds (UCB), which minimizes the cumulative regret at each node \citep{AuerCF02}.
Theoretical bandits were soon replaced with empirical bandits, giving
better results. First the RAVE algorithm \citep{Gelly2011AI} improved
greatly on UCT for the games of Go and Hex \citep{cazenave2009hex}.
A later refinement is GRAVE that improves on RAVE for many different
games \citep{cazenave2015grave} and is used for General Game Playing,
for example in Ludii \citep{browne2019practical}.
MCTS was combined with neural networks in AlphaGo, surpassing professional
level in the game of Go \citep{silver2016mastering}. The search algorithm
used in AlphaGo is MCTS with PUCT, a bandit that uses the policy given
by the neural network to bias the moves to explore. Later AlphaGo
was redesigned to learn from zero knowledge, leading to AlphaGo Zero
\citep{silver2017mastering}. It was then applied to other games, namely
Shogi and Chess, with the more general AlphaZero program \citep{silver2018general}.
Many teams have replicated the AlphaZero approach for Go and for other
games: Elf/OpenGo \citep{tian2019elf}, Leela Zero \citep{pascutto2017leela},
Crazy Zero by Rémi Coulom, KataGo \citep{wu2019accelerating}, Galvanise
Zero \citep{gzero}, and Polygames \citep{cazenave2020polygames}.

As we use Polygames as a sparring partner, we will give more details
about it. Polygames (MIT License) replicates the AlphaZero approach
and has been successfully applied to many games. There are multiple innovations
in Polygames. It can train neural networks with an architecture independent
of the size of the board. To do so it uses a fully convolutional policy,
meaning that there is no dense layer between the last convolutional
planes and the policy. The value head is also independent of the size
of the board since it uses global average pooling before the dense
layers connected to the evaluation output neuron. It is much more
difficult to train a network for Hex $19$ (board size $19\times19$)
or Havannah $10$ than training it on a smaller size board. Polygames
did succeed in these games by scaling its neural networks trained
on smaller sizes to the difficult board sizes. It played on the Little
Golem game server and beat the best players at these games that were
considered too difficult for \emph{Zero Learning} (i.e. learning without using knowledge except the game rule). Other innovations of Polygames
include a pool of neural networks during self-play matches in order to avoid catastrophic
forgetting.

The other kinds of algorithms used in computer games are the $\alpha\beta$
family of algorithms, whose apogee took place with Deep Blue
\citep{campbell2002deep}, the first program beating a world chess
champion. $\alpha\beta$ dominated the field of perfect information
games until the advent of MCTS in 2006. Still, many current strong
Chess programs use $\alpha\beta$ \citep{haworth202120}. The latest
versions are combined with NNUE neural networks \citep{nasu2018efficiently}.
There has been a lot of research on the optimizations of $\alpha\beta$
 \citep{marsland1987computer}. Many of them deal with move
ordering since move ordering can drastically improve the search time
of $\alpha\beta$ \citep{knuth1975}. In \cite{takada2019reinforcement}, $\alpha\beta$ 
 was also combined with a policy within a reinforcement learning architecture and it reaches a good level at Hex (the policy is used to prune actions in order to reduce the branching factor).

The search algorithms we use to learn and play games are close to Unbounded
Best-first Minimax Search \citep{korf1996best}. There is very little
study on this algorithm and it seems little or not applied in practice,
except in the work of \citep{cohen2020learning}. In that work, variants
and improvements of Unbounded Minimax are proposed with several complementary
techniques of zero learning that do not require the use of policies. The proposed overall architecture, called \emph{Athénan} (or also \emph{the Descent framework}), exceeds the state-of-the-art level
of play at the game of Hex (size $11$, $13$, and $19$) and other games. In the
context of the experiments of \citep{cohen2020learning}, Athénan is the best zero learning approach not using a policy: in
particular, replacing the used variant of Unbounded Minimax, called
\emph{Descent Minimax} (or Descent for short), by Unbounded Minimax, by $\alpha\beta$, or by MCTS
(with UCT) gives less good results. Moreover, in the experiments of
that work, another variant of Unbounded Minimax, called \emph{Unbounded
Minimax with Safe decision}, is shown better than Unbounded Minimax
and than $\alpha\beta$ for confrontations (``What is the best search
algorithm for winning a game?'' is a different question than ``What
is the best search algorithm to learn faster?''). 
In \citep{cohen2021completeness}, it has been proved that, with enough time, Descent Minimax and Unbounded Minimax find the best game strategy (multiplayer generalizations are also proposed). 

\begin{algorithm}[!bh]
\DontPrintSemicolon\SetAlgoNoEnd

\SetKwFunction{descenteiteration}{descent\_iter}

\SetKwFunction{descente}{descent}\SetKwFunction{time}{time}\SetKwFunction{actions}{actions}\SetKwFunction{bestaction}{best\_action}\SetKwFunction{terminal}{terminal}\SetKwFunction{premier}{first\_player}
\SetKwProg{myproc}{Function}{}{}

\myproc{\descenteiteration{$s$, $S$, $T$, $\hadapt$, $\hterminal$}}{

\eIf{\terminal{$s$}}{

$S\leftarrow S\cup\{s\}$\;

$v(s)\leftarrow \hterminal(s)$

}{

\If{$s\notin S$}{

$S\leftarrow S\cup\{s\}$\;

\ForEach{$a\in$ \actions{$s$}}{

\eIf{\terminal{$a(s)$}}{

$S\leftarrow S\cup\{a(s)\}$\;

$v(s,a)\leftarrow \hterminal\left(a(s)\right)$\;

$v(a(s))\leftarrow v(s,a)$\;

}{

$v(s,a)\leftarrow \hadapt\left(a(s)\right)$\;

}

}}

$a_{b}\leftarrow$ \bestaction{$s$}\;

$v(s,a_{b})\leftarrow$ \descenteiteration{$a_{b}(s)$, $S$, $T$, $\hadapt$, $\hterminal$}\;

$a_{b}\leftarrow$ \bestaction{$s$}\;

$v(s)\leftarrow v(s,a_{b})$\;

}

return $v(s)$\;

}

\;

\myproc{\descente{$s$, $S$, $T$, $\hadapt$, $\hterminal$, $\tau$}}{

$t=$ \time{}\;

\lWhile{\time{}$-\,t<\tau$}{\descenteiteration{$s$, $S$, $T$, $\hadapt$, $\hterminal$}}

return $S$\;

}\;

\protect\protect

\caption{\emph{Descent} algorithm (symbol definitions in Table~\ref{tab:Index-of-symbols}).\label{alg:descente}}
\end{algorithm}

\subsection{Deep Reinforcement Learning Algorithms Compared in this Paper\label{learning}}

We detail in this section the two zero learning frameworks used in
the experiments of this article.

\begin{table}[t]
\begin{centering}
{\footnotesize{}}%
\begin{tabular}{|c|c|}
\hline 
{\footnotesize{}Symbols} & {\footnotesize{}Definition}\tabularnewline
\hline 
\hline 
{\footnotesize{}$\mathrm{actions}\left(s\right)$} & {\footnotesize{}action set of the state $s$ for the current player}\tabularnewline

\hline 
{\footnotesize{}$\mathrm{terminal\left(s\right)}$} & {\footnotesize{}true if $s$ is an end-game state}\tabularnewline
\hline 
{\footnotesize{}$a(s)$} & {\footnotesize{}state obtained after playing the action $a$ in the
state $s$}\tabularnewline
\hline 
{\footnotesize{}$\mathrm{time}\left(\right)$} & {\footnotesize{}current time in seconds}\tabularnewline
\hline 


    \multirow{2}{*}{ \footnotesize{}$S$ }
    & {\footnotesize{}states of the partial game tree} \tabularnewline
    & {\footnotesize{}
 (and keys of the transposition table $T$)}
\tabularnewline
\hline 

{\footnotesize{}$T$} & {\footnotesize{}transposition table (contains state labels as $v$ or $P$)}\tabularnewline
\hline 
{\footnotesize{}$P(s)$} & {\footnotesize{} target policy of state $s$ computed from the search data}\tabularnewline

\hline 
{\footnotesize{}$D$} & {\footnotesize{}learning data set}\tabularnewline
\hline 
{\footnotesize{}$\tau$} & {\footnotesize{}search time per action}\tabularnewline

\hline 
{\footnotesize{}$t_{\max}$} & {\footnotesize{}chosen total duration of the learning process }\tabularnewline

\hline 
{\footnotesize{}$v(s)$} & {\footnotesize{}value of state $s$ from the game search}\tabularnewline
\hline 
{\footnotesize{}$v(s,a)$} & {\footnotesize{}value obtained after playing action $a$ in state
$s$}\tabularnewline
\hline

    \multirow{2}{*}{ \footnotesize{}$\hadapt(s)$ }
    & {\footnotesize{}adaptive evaluation function (of non-terminal game} \tabularnewline
    & {\footnotesize{}
tree leaves ; first player point of view)}
\tabularnewline
\hline

    \multirow{2}{*}{ \footnotesize{}$\hterminal(s)$ }
    & {\footnotesize{}evaluation of terminal states, e.g. } \tabularnewline
    & {\footnotesize{}
 game gain  (first
player point of view)}
\tabularnewline
\hline 

    \multirow{2}{*}{ \footnotesize{}action\_selection($s$, $S$, $T$) }
    & {\footnotesize{}decides the action to play in the state $s$ } \tabularnewline
    & {\footnotesize{}
depending
on the partial game tree, i.e. on $S$ and $T$}
\tabularnewline
\hline 

    \multirow{2}{*}{ \footnotesize{}update($\hadapt,D$) }
    & {\footnotesize{}updates the parameter $\theta$ of $\hadapt$ in order} \tabularnewline
    & {\footnotesize{}
for $\hadapt(s)$ is closer to $v$ for each $(s,v)\in D$}
\tabularnewline
\hline 
\end{tabular}{\footnotesize\par}
\par\end{centering}
\caption{Index of symbols\label{tab:Index-of-symbols}}
\end{table}

\subsubsection{Polygames Learning Algorithm\label{subsec:Polygames_learning}}

Polygames uses its search algorithm, MCTS with PUCT, to generate matches,
by playing against itself. It uses the information from these matches
to update its neural network. This neural network is used by the search
algorithm to evaluate states by a value and by a policy (i.e. a probability
distribution on the actions playable in that state). For each finished
match, the network is trained to associate with each state of the
state sequence of this match the result of the end of that match (which
is $-1$ for a loss,  $0$ for a draw, and $+ 1$ for a win). It is also trained, at the same time, to associate
with each state a particular policy. In that ``target'' policy, the probability of an action
 is proportional to $N^{\nicefrac{1}{\tau}}$
where $N$ is the number of times this action has been selected in
the search from that state and $\tau$ is a parameter. Note that,
during each Polygames learning process, several games are performed
in parallel and their evaluations are batched in order to be evaluated
in parallel on the GPU.

\subsection{More Details about AlphaZero/Polygames}
During a learning process using AlphaZero (and thus Polygames), as long as there is time left, a new match phase is performed. A phase consists of a match against oneself, where in each turn the move to be played is decided after carrying out a search with MCTS + PUCT. MCTS is similar to Unbounded Minimax. The first main difference is that the value of a state is not the minimax value in the partial game tree but the average of the leaves in the subtree starting from that state. The second main difference is that the tree is constructed not in choosing states of higher value, but states optimizing the value plus an exploration term depending on the policy of the neural network and the number of selection of actions during the search. After the match, the match state sequence data is added to the previous data (only the most recent data points are kept). Periodically, 
training is performed from a sample of this data set. The main part of this algorithm is described in Algorithm \ref{alg:polygames}.

\begin{algorithm}[!bh]
\DontPrintSemicolon\SetAlgoNoEnd

\SetKwFunction{terminalearning}{AlphaZero\_main\_algorithm}\SetKwFunction{search}{search}\SetKwFunction{mcts}{mcts}\SetKwFunction{initial}{initial\_game\_state}\SetKwFunction{processing}{processing}\SetKwFunction{update}{update}

\SetKwFunction{actionselection}{action\_selection}\SetKwFunction{terminal}{terminal}
\SetKwProg{myproc}{Function}{}{}

\myproc{\terminalearning{$t_{\max}$, $\tau$}}{

$t_{0}\leftarrow$ \time{}\;

\While{\time{}$-\,t_{0}<t_{\max}$ }{

\For{$k\in \{1,\ldots,K\}$}{

$s\leftarrow$\initial{}\;

$S\leftarrow\emptyset$\;

$T\leftarrow\{\}$\;

$G\leftarrow\left\{ s\right\} $\;

\While{$\neg$\terminal{$s$}}{

$S,\,T\leftarrow$\mcts{$s$, $S$, $T$, $\hadapt$, $\hterminal$, $\tau$}\;

$a\leftarrow$\actionselection{$s$, $S$, $T$}\;

$s\leftarrow a(s)$\;

$G\leftarrow G\cup\left\{ s\right\} $\;

}\;

$D\leftarrow\{\left(s', \left(\hterminal(s), P(s')\right)\right)\ |\ s'\in G\}$\;


}\;

\update{$\hadapt$, $D$}\;

}\;

}\;

\protect\protect

\caption{Main algorithm of AlphaZero (see Table~\ref{tab:Index-of-symbols}
    for the definitions of symbols ; $K$ is the number of matches performed between two updates, some of these matches are executed in parallel ; $G$ is the sequence of states of the current match).\label{alg:polygames}}
\end{algorithm}

\subsubsection{Athénan Standard Learning Algorithm\label{subsec:descent}}

Athénan, the learning framework of \citep{cohen2020learning}, is based on a
variant of Unbounded Minimax called \emph{Descent Minimax} (or \emph{Descent} for short), 
which consists in exploring the sequences of actions until terminal
states. In comparison, Unbounded Minimax and MCTS explore the sequences
of actions only until reaching a leaf state. An iteration of Descent (an analyzed sequence of actions)
thus consists in a deterministic complete simulation of the rest of
the game. The exploration is thus deeper while remaining a best-first
approach. This allows the values of terminal states to be propagated
more quickly to (shallower) non-terminal states. Descent is formally
described in Algorithm~\ref{alg:descente}.

Unlike Polygames, the learned target value of a state is not the end-game
value but its minimax value in the partial game tree built during
the match. This information is more informative, since it directly contains
part of the knowledge acquired during the previous matches. In addition,
contrary to Polygames, learning is carried out for each state of the
partial game tree constructed during the searches of the match (not
just for each state of the states sequence of the played match). In
other words, with Polygames, there is one learning target per search
whereas with Athénan, there are several learning targets
per search. Therefore, there is no
loss of information with Athénan: all of the information acquired during the search
is used during the learning process. As a result, Athénan
generates a much larger amount of data for training from the same
number of played matches than AlphaZero / Polygames. Thus, unlike
the state of the art which requires to generate matches in parallel
to build its learning dataset, this approach does not require the
parallelization of matches (and the parallelization of Athénan is
not done in the experiments of this article).

During confrontations, the used search algorithm is Unbounded (Best-First)
Minimax with Safe decision, denoted $\ubfms$. It is a variant of
Unbounded Best-First Minimax which performs the same search.
More precisely, it iteratively extends the best sequence of actions
in the partial game tree (i.e. it adds at each iteration the leafs
of the principal variation of the partial game tree). Note that, on
the one hand, the best action sequence generally changes after each
extension. On the other hand, in general, the worse the evaluation
function is, the wider the exploration is. The difference between them is as follows: with Unbounded
Minimax, the action to play, chosen after each search, is the one with the best value, while with this variant, the chosen action  is the one that is the most
  explored.

Finally, this approach is optionally based on a \emph{reinforcement
heuristic}, that is to say an evaluation function of terminal states
more expressive than the classical gain of a game (i.e. $+1$ / $0$
/ $-1$). The best proposed general reinforcement heuristics in \citep{cohen2020learning}
are \emph{scoring} and the \emph{depth heuristic} (the latter favoring
quick wins and slow defeats). 

Note that this approach does not use a policy, so there is no need
to encode actions. Consequently, this avoids the learning performance
problem of neural networks for games with large number of actions
(i.e. very large output size). In addition, although Athénan
does not performed matches in parallel, it batches all the child states
of an extended state together to be evaluated at one time on the GPU \citep{cohen2020learning} (
with Descent and Unbounded Minimax).

\subsection{More Details about Athénan}
During a learning process using Athénan, as long as there is time left, a new match phase is performed. A match phase consists of a match against oneself, where in each turn the move to be played is decided after carrying out a search with Descent. The move to be played after the search is chosen according to an action selection method, depending on the result of the search. In these experiments, the used action selection method is the \emph{ordinal law} (actions are chosen randomly according to the order of their value) \cite{cohen2020learning} with the exploitation parameter $\epsilon'$ chosen at random uniformly between 0 and 1 each time a new action must be decided. After each match phase, the data from the associated partial game tree is added to the previous data (here, only the data of the last 100 matches are kept). Then, a training phase is carried out from a sample of this data set. 
Specifically, \emph{smooth experience replay} is used \cite{cohen2020learning}. The main part of this algorithm is described in Algorithm \ref{alg:cadre_de_descente}.  The full formalization is described in \cite{cohen2020learning}.

\begin{algorithm}[!bh]
\DontPrintSemicolon\SetAlgoNoEnd

\SetKwFunction{iterationUM}{UBFM\_iteration}
\SetKwFunction{UM}{UBFM}\SetKwFunction{time}{time}\SetKwFunction{actions}{actions}\SetKwFunction{bestaction}{best\_action}\SetKwFunction{terminal}{terminal}\SetKwFunction{premier}{first\_player} \SetKwProg{myproc}{Function}{}{}

\SetKwFunction{treelearning}{Athénan\_main\_algorithm}\SetKwFunction{descent}{descent}\SetKwFunction{search}{search}\SetKwFunction{initial}{initial\_game\_state}\SetKwFunction{learningtime}{learning\_time}\SetKwFunction{processing}{processing}\SetKwFunction{update}{update}

\SetKwFunction{actionselection}{action\_selection}\SetKwFunction{terminal}{terminal}\SetKwFunction{leaf}{leaf}
\SetKwProg{myproc}{Function}{}{}

\myproc{\treelearning{$t_{\max}$, $\tau$}}{

$t_{0}\leftarrow$ \time{}\;

\While{\time{}$-\,t_{0}<t_{\max}$ }{

$s\leftarrow$\initial{}\;

$S\leftarrow\emptyset$\;

$T\leftarrow\{\}$\;

\While{$\neg$\terminal{$s$}}{

$S,\,T\leftarrow$\descent{$s$, $S$, $T$, $\hadapt$, $\hterminal$, $\tau$}\;

$a\leftarrow$\actionselection{$s$, $S$, $T$}\;

$s\leftarrow a(s)$

}\;

$D\leftarrow\{\left(s,v(s)\right)\ |\ s\in S\}$\;


\update{$\hadapt$, $D$}

}\;

}\;

\protect\protect

\caption{Main algorithm of Athénan (see Table~\ref{tab:Index-of-symbols}
for the definitions of symbols).\label{alg:cadre_de_descente}}
\end{algorithm}

\section{Comparison of Zero Reinforcement Learning Algorithms\label{sec:compar_learning}\label{sec:compar_learning2}}

In this section, we experimentally compare the two learning algorithms
Polygames (see Section \ref{subsec:Polygames_learning})
and Athénan (see Section \ref{subsec:descent}). First, in the context
of 8 games, we compare the data efficiency of the two algorithms,
i.e. the amount of data generated during the self-play matches which
are learned in order to self-improve. Second, we compare the win performances
of the two algorithms in the same context (in particular, the algorithms 
 use the same resources). They are rated against MCTS.
Then, a longer training is performed
on Hex 13 and the algorithms are evaluated against Mohex
2.0 \citep{huang2013mohex}, the best publicly available Hex program.
Finally, the Polygames networks, that have won numerous medals during
the TCGA 2020 tournament, confront Athénan networks that have used
drastically less computational power for their learning processes. In each of these experiments,
Athénan is strongly better than Polygames.

\subsection{Technical Details}

We expose in this section the technical details common to the experiments of Sections \ref{data_comparison}, \ref{win_comparison_same}, and \ref{long_win_comparison_same}. Recall that full details of experiments of this paper are in Technical Appendix \cite{cohen2023appendix}.

\subsubsection{Parameters}

 For each learning process with Athénan, the batch size of the stochastic gradient descent $B$ is 3000, \emph{smooth experience replay} is used with the following parameters: $\mu=100$ and $\delta=3$.
The neural architecture is the same for each game: a $R_2$-network (see Table \ref{tab:architecture}).
The number of parameters in each neural network is of the order of $5\cdot 10^6$. This implies that the number of filters $F$ and number of dense neurons $D$ are different for each game. The corresponding numbers are described in Table \ref{tab:weights_descent}.

The action distribution used during the learning process is the ordinal law \cite{cohen2020learning}. It is used with a uniform random variable between 0 and 1 as exploration parameter (the variable value changes after each search performed for determining the next action to play; therefore no simulated annealing is used).

\begin{table}
\centering{}%
\begin{tabular}{|c|c|c|c|}
\hline 
layer $\#$ & $C$-network & $R_{1}$-network & $R_{2}$-network\tabularnewline
\hline 
\hline 
$1$ & conv. + ReLU & convolution & convolution\tabularnewline
\hline 
$\cdots$ & conv. + ReLU & $2$ res. blocks & $8$ res. blocks\tabularnewline
\hline 
$N-2$ & conv. + ReLU & $1\times1$ conv. & dense + ReLU\tabularnewline
\hline 
$N-1$ & dense + ReLU & dense + ReLU & dense + ReLU\tabularnewline
\hline 
$N$ & dense layer & dense layer & dense layer\tabularnewline
\hline 
\end{tabular}\caption{Description of $3$ neural architectures of value networks, called
$C$-network, $R_{1}$-network, and $R_{2}$-network. Each residual
block is composed of a ReLU followed by a convolution followed by
a ReLU followed by a convolution. Output contains
one neuron. Other parameters are: kernel is $3\times3$, filter number
is $F$, number of neurons in dense layers is $D$,  padding is used
with $R_{i}$-network but not with $C$-network. \label{tab:architecture}}
\end{table}

\begin{table}
\begin{centering}
\begin{tabular}{|c|c|c|c|c|c|}
\hline 
Game & $F$ & $D$ & Game & $F$ & $D$\tabularnewline
\hline 
\hline 
Surakarta & 132 & 845 & Breakthrough & $132c$ & 477\tabularnewline
\hline 
Othello & 132 & 477 & Outer-Open-Gomoku & 132 & 111\tabularnewline
\hline 
Hex $13$ & 132 & 155 & Havannah 8 & 132 & 111\tabularnewline
\hline 
Connect6 & 132 & 65 & Havannah 10 & 132 & 65\tabularnewline
\hline 
\end{tabular}

\par\end{centering}
\caption{The filter number in convolutional layers and the number of neurons
in dense layers of the $R_{2}$-networks used with
Athénan, detailed for the $8$ games.\label{tab:weights_descent}}
\end{table}

Network architectures used for Polygames are adaptations of the architecture being used with Athénan, in order to add a policy while keeping an analogous number of parameters in the neural network (see the Supplementaries document for the details).

Evaluations of Section \ref{win_comparison_same} are performed  against the basic MCTS algorithm based on UCT (it uses 160 rollouts). For each learning process (i.e. each learned neural network), each evaluation consists of 400 games (200 in first player and 200 in second player). 

\subsubsection{Computational Resources\label{subsect:ressources}}
 In this section, we present the used computational resources for the experiments of this paper.
 
For the performed training runs and confrontations, we use the following hardware: GPU Nvidia Tesla V100 SXM2 32 Go, $2$ to $10$ CPU (processors Intel Cascade Lake 6248 2.5GHz) on RedHat.
There is an exception, for the performed confrontations against Polygames tournament networks (confrontations of Section \ref{subsec:tournament-comparison}), we use the following hardware:  GeForce GTX 1080 Ti, $2$ to $8$ CPU (Intel(R) Xeon(R) CPU E5-2603 v3 1.60GHz) on Ubuntu 18.04.5 LTS.

Athénan programs (Descent Minimax, Unbounded Minimax, ...)
are coded in Python (using tensorflow 1.15). Games and Search in Polygames are coded in C/C++. For confrontations, Polygames num\_actor parameter is $8$ 
(threads doing MCTS).

\subsection{Comparison of Generated Learning Data\label{data_comparison}}

In this section, we experimentally compare the number of state data,
the number of state evaluations, and the number of neural network evaluations
performed during an Athénan training and a Polygames training, each during 15 days. In total, 8 trainings were carried out with Athénan and
5 with Polygames for each of the following games: Connect6, Outer-Open-Gomoku,
Hex 13, Havannah 8, Havannah 10, Othello, Breakthrough, and Surakarta. 

We start by comparing the number of evaluations. In
average, the neural network evaluations of Athénan is 12.7 times smaller
than that of Polygames. In addition, the average number of state evaluations
of Athénan is 2.5 times smaller than that of Polygames. In other words,
Polygames is more efficient to perform evaluations. However, this
is not an intrinsic characteristic, because this difference is mainly
explained by two facts. First, Athénan is coded in Python but searches
and game mechanisms for Polygames are coded in C/C++, which allows
it to be 2 to 5 times faster (the speed difference depends on the game ; for example in Python, at Othello, game calculations take more than $82\%$ of the time of the learning process). Second, Polygames
performs many matches in parallel, whereas Athénan, in its implementation,
is purely sequential (except for evaluating the children of a state: these evaluations are simultaneously performed on the GPU). Thus, the matches parallelization of Polygames gives a potentially larger number of evaluations for
the same period of time (Polygames evaluations are also simultaneously performed on the GPU). The detailed numbers for each game are described
in Table \ref{data_stats}. In summary, in this experiment, Polygames
performs more evaluations but it could be counterbalanced by implementing
Athénan in C/C++ or by parallelizing it.

Now we compare \emph{learned states}: the number of state data used
during the learning process.  Athénan generates 296 times
more learned states than Polygames (despite performing fewer
evaluations as we saw in the previous paragraph). This is due to the
fact that Athénan uses tree learning: it learns all the data generated
during the search, i.e. it learns all the partial game tree build during the search. By contrast, Polygames / AlphaZero only learns a
summary of this search, namely a policy and a state value for the
state analyzed during the search. Note that since determining a policy
for a state requires that its children be sufficiently explored, it
does not seem possible to learn a policy for each state of the search (i.e. for each state of the partial game tree). In other words,
it
does not seem possible to perform tree learning for the policy with AlphaZero / Polygames.
The same remark applies for the state value. Indeed, since the learning
target for the state value is the endgame value, it would be the
same for all the states of the tree, which is more likely to negatively
impact the training in creating over-fitting than improving learning.
In other words, naively modifying AlphaZero / Polygames to use a terminal tree learning and tree
learning for the policy in order to decrease
the cost of data generation should not improve performance. However,
it is possible to change the learning target, i.e. replace the endgame value by the search state value and thus to perform classic tree learning. This has
been studied in the context of MCTS without policy, and the
results are much worse than with Descent Minimax and tree learning (i.e. with Athénan standard algorithms) \citep{cohen2020learning}.


In conclusion, the cost of state data generation is 50 to 700 times
better with Athénan than with Polygames depending on the game (296
times better on average over the tested games), despite the fact that
it performs 2.5 times fewer states evaluations. 
Moreover,
recall that using a language other than Python with Athénan would
further improve its performance, most likely by a factor of at least
$2$. Note also that the matches with Athénan are not parallelized
(unlike AlphaZero / Polygames), and parallelizing them would also
increase the number of data.

\subsection{Win comparison with Same Resources\label{win_comparison_same}}

In this section, we compare the learning performances of Athénan
 with the learning performances of Polygames, with respect
to the win percentages against MCTS. 

This comparison is notably based on the \emph{gain} of using Athénan rather than Polygames, which is the difference in their win performances:

\begin{defn} \label{def:gain}
The \emph{gain of using the algorithm $A$ rather than the algorithm $A'$} is ${\frac{1}{2}\left((w_{A}-l_{A})-(w_{A'}-l_{A'})\right)}$
where $w_{A}$ (resp. $l_{A}$) is the win (resp. loss) percentage
of $A$.


\end{defn}

Several trainings, each during
15 days, have been performed for each of the following games: Surakarta,
Hex $13$, Connect6, Outer-Open-Gomoku, Breakthrough, Othello $8$,
Havannah $8$, Havannah $10$. In total, 5 repetitions were performed
with Polygames and 8 repetitions were performed with Athénan for each
game (4 repetitions without reinforcement heuristic and 4 repetitions
with reinforcement heuristic). As the number of repetitions is small,
we use the following advanced statistical evaluation procedure: \emph{stratified
bootstrap confidence interval} \citep{agarwal2021deep} which allows
one to evaluate learning processes over several tasks even with a
low number of repetitions. 

\begin{figure}
\begin{centering}

{\includegraphics[scale=0.34]{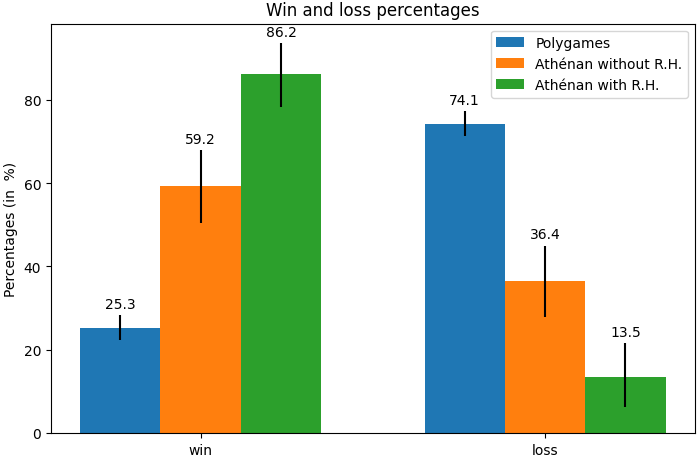}}

\caption{Performance of Athénan (resp. Polygames) against MCTS with UCT at the end of the $15$ days of training averaged over the $8$ games.
Their stratified bootstrap
confidence intervals are indicated by the black lines. 
Athénan results are detailed in function of the use of a reinforcement heuristic (abbreviated R.H.). \label{fig:res_descent}}

\end{centering}
\end{figure}

\begin{figure*}[htb]
\begin{centering}

{\includegraphics[scale=0.395]{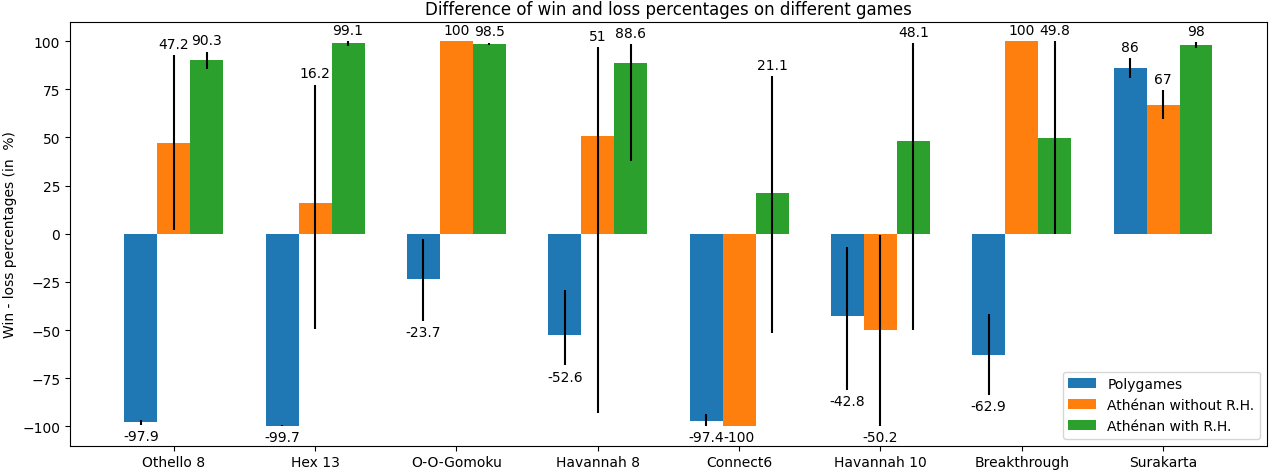}}

\caption{Average win percentages minus loss percentages of Athénan (resp. Polygames)
against MCTS with UCT at the end of the $15$ days of training for the 8 games. 
Athénan results are detailed in function of the use of a reinforcement heuristic (abbreviated R.H.). Their bootstrap
confidence intervals are indicated by the black lines.\label{fig:res_descent-1}}

\end{centering}
\end{figure*}

\begin{figure}
\begin{centering}
{\includegraphics[scale=0.265]{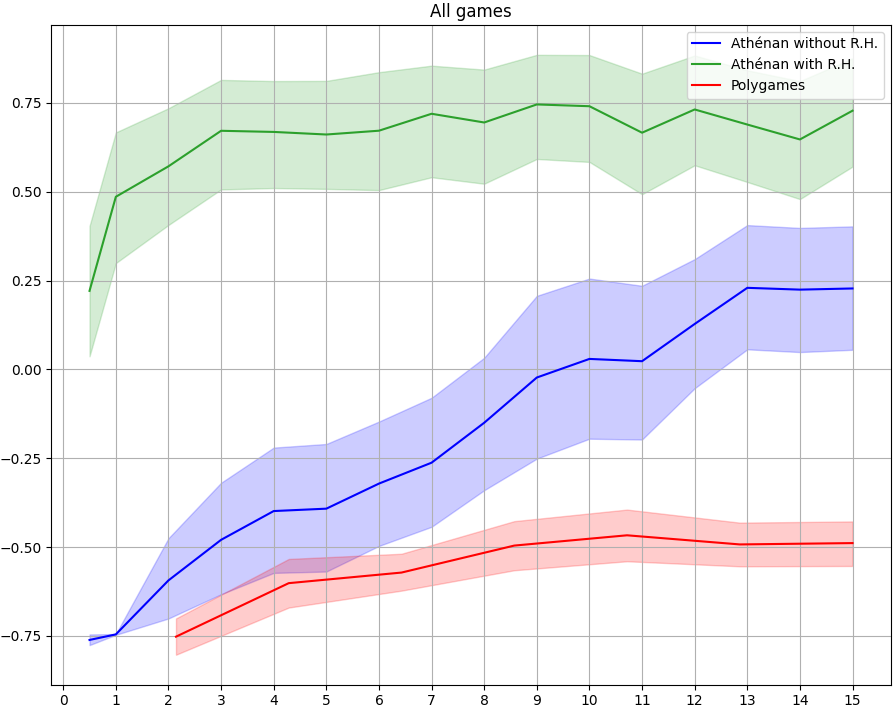}}
{\caption{Evolution of average win rates minus average loss rates of Athénan with reinforcement heuristic (with R.H.),  of Athénan without reinforcement heuristic, and of Polygames against MCTS
with UCT along the 15 days of training and their stratified bootstrap
confidence intervals over the $8$ games. \label{fig:courbe}}}

\end{centering}
\end{figure}

The final global performances of the learning processes based on Athénan and Polygames against a 160-rollouts MCTS with UCT (i.e. without
any knowledge nor learned policy) are described in Figure \ref{fig:res_descent}. The details for each of the $8$ games are described  in Figure~\ref{fig:res_descent-1}.
The curves
describing the evolution of the performances of the two algorithms throughout the training
are described in Figure~\ref{fig:courbe}.

In conclusion, the performances of Athénan are much better than those
of Polygames. The performance superiority of Athénan is even more marked when a reinforcement heuristic
is used (we already knew that reinforcement heuristic is an improvement
of the combination of Descent Minimax and tree learning, i.e. an essential component of Athénan \citep{cohen2020learning}). In particular, on the one hand, the gain of
using Athénan without reinforcement heuristic rather than Polygames
is $35.8\%$ (see Def. \ref{def:gain}). On the other hand, the gain of using Athénan with reinforcement heuristic
rather than Polygames is $60.75\%$. Regarding learning speed, Athénan without
reinforcement heuristic achieves in only $2$ days the performance of Polygames after 15 days of training, i.e. there is a  factor of $7$. Moreover, Athénan with reinforcement heuristic
achieves this performance in much less than half a day, there is a factor of $30$ (see the curve in
Figure~\ref{fig:courbe}).

\subsection{Win Comparison with Same Resources during a Long-Term Learning Process \label{long_win_comparison_same}}

In this section, we compare again the learning performances of Athénan with the learning performances of Polygames with
respect to win percentages, but for a longer training (113 days),
and only at Hex 13, evaluating them this time against Mohex 2.0 \citep{huang2013mohex}, champion
program at Hex from 2013 to 2017 at the Computer Olympiads. Mohex 2.0 is the strongest
hex program which is freely available. 

For this, we have continued the learning processes of the previous section carried out on Hex $13$ with Athénan and with Polygames. Then, we have thus evaluated them against Mohex 2.0, at different stages of their learning processes (an evaluation has been performed approximately every 4 days).

\begin{figure}[htb]
\begin{centering}
\includegraphics[scale=0.265]{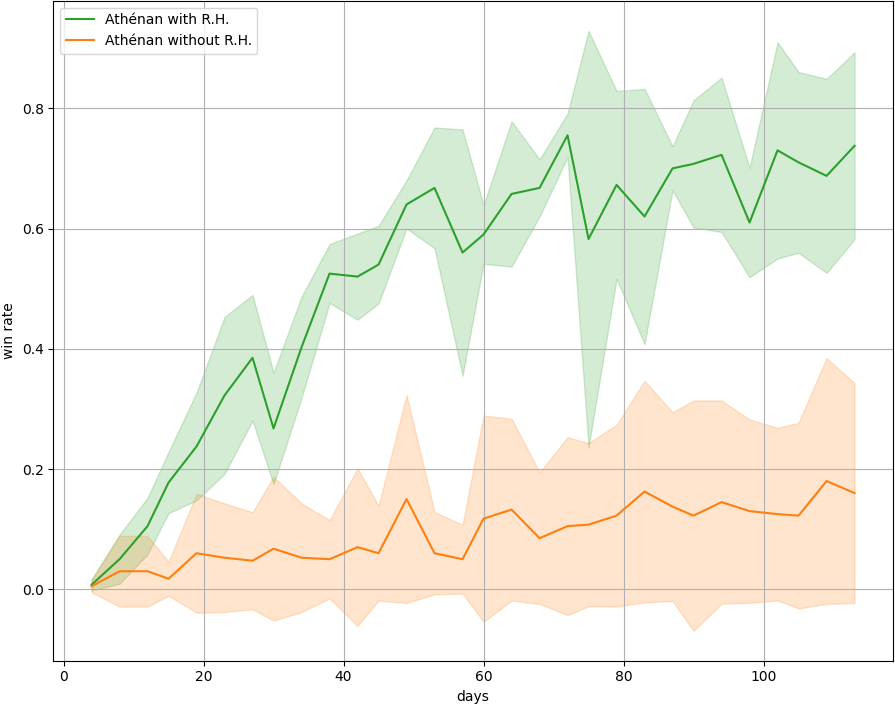}
\par\end{centering}
\caption{Evolution of average win rates of Athénan with and without reinforcement heuristic (R.H.)
 against Mohex 2.0, during 113 days of training
(there is approximately one evaluation every 4 days ; each
evaluation consists of 50 matches in first player and 50 other matches
in second player). 
Shading is the $95\%$ confidence interval.\label{fig:hex_learning}}
\end{figure}

The evolution of the average win percentages of Athénan (with and without reinforcement heuristic) 
against Mohex 2.0  during the training  
 is shown in Figure \ref{fig:hex_learning}. Athénan with reinforcement heuristic goes rather far beyond the level of Mohex 2.0. Athénan without reinforcement heuristic does not reach the level of Mohex 2.0 but it still manages to beat it in certain positions.
 On the contrary, none of the learned Polygames networks (combined with the Polygames search algorithm) has succeeded in winning any match against Mohex 2.0, whatever the evaluation moment during their learning process. In other words,  the Polygames winning curve is constant and is $0\%$, with a confidence interval of $[0\%;0\%]$.

Therefore, in this experiment, learning with \emph{Athénan} is also widely better
than with Polygames. 

\subsection{Comparison versus Tournaments Polygames Networks\label{subsec:tournament-comparison}}

In this section, we evaluate Athénan networks against high level Polygames
networks, at Breakthrough, at Othello $8$ and $10$. 

The Polygames networks are those having won at Breakthrough and Othello
$10$ and finished second at Othello $8$ in the TCGA 2020 tournament.
They have been trained during 7 days with 100 GPU each. The used Athénan
networks was trained with only one GPU during $5$ days. These trainings was later extended to $30$ days.

\begin{figure}
\begin{centering}
\includegraphics[scale=0.22]{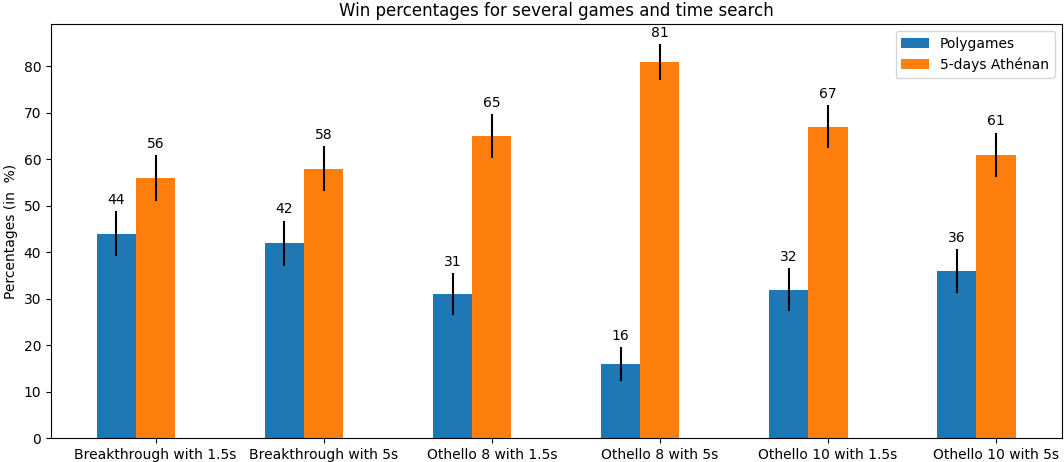}
\par\end{centering}
\caption{Results of $400$ matches between Athénan ($5$ days of training) and Polygames (using tournaments Polygames networks) at Breakthrough, Othello
$8$, and Othello $10$. \label{fig:vs_tournament_day5}}
\end{figure}

The results of the confrontation of the 5-day Athénan networks against
Polygames networks are described in Figure \ref{fig:vs_tournament_day5}.
Although learning with Athénan required $100$ times less GPU ($1$ GPU
vs. $100$ GPU) and lasted slightly less time (5 days vs. a week),
the performance of Athénan is much better for each of the three games.

\begin{figure}
\begin{centering}
\includegraphics[scale=0.21]{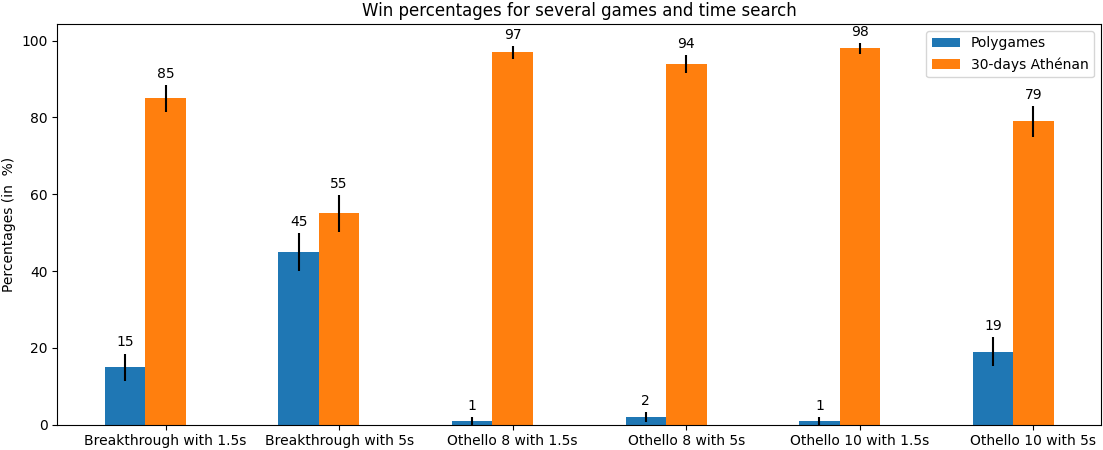}
\par\end{centering}
\caption{Results of $400$ matches between Athénan ($30$ days of training) and Polygames (using tournaments Polygames networks)  at Breakthrough, Othello
$8$, and Othello $10$.  \label{fig:vs_tournament_day30}}
\end{figure}

The results of the confrontation of the 30-day Athénan networks against
the same Polygames networks are described in Figure \ref{fig:vs_tournament_day30}.
This 30 days experience shows that the Athénan networks continue to improve.

\begin{table*}[!th]
\begin{centering}
{\footnotesize{}{}}{\scriptsize{}}%
\begin{tabular}{|c|c|c|c|c|c|c|c|c|}
\hline 
 & {\scriptsize{}{}Connect6} & {\scriptsize{}{}Havannah $10$} & {\scriptsize{}{}Havannah $8$} & {\scriptsize{}{}Outer-Open-Gomoku} & {\scriptsize{}{}Hex $13$} & {\scriptsize{}{}Surakarta} & {\scriptsize{}{}Othello} & {\scriptsize{}{}Breakthrough}\tabularnewline
\hline 
\hline 
{\scriptsize{}{}Learned states} & {\scriptsize{}{}55} & {\scriptsize{}{}64} & {\scriptsize{}{}111} & {\scriptsize{}{}115} & {\scriptsize{}{}359} & {\scriptsize{}{}442} & {\scriptsize{}{}529} & {\scriptsize{}{}693}\tabularnewline
\hline 
{\scriptsize{}{}Neural evaluation} & {\scriptsize{}{}0,02} & {\scriptsize{}{}0,03} & {\scriptsize{}{}0,05} & {\scriptsize{}{}0,04} & {\scriptsize{}{}0,10} & {\scriptsize{}{}0,11} & {\scriptsize{}{}0,12} & {\scriptsize{}{}0,16}\tabularnewline
\hline 
{\scriptsize{}{}States evaluation} & {\scriptsize{}{}0,37} & {\scriptsize{}{}0,30} & {\scriptsize{}{}0,37} & {\scriptsize{}{}0,49} & {\scriptsize{}{}0,65} & {\scriptsize{}{}0,40} & {\scriptsize{}{}0,10} & {\scriptsize{}{}0,49}\tabularnewline
\hline 
\end{tabular}{\scriptsize\par}
\par\end{centering}
{\footnotesize{}{}\caption{Ratio of Athénan data over Polygames data for the same learning time
and for different games (average over 5 runs for Polygames and 8 runs for Athénan ;
data of a run varies by a maximum of $\pm60\%$ for Polygames and
$\pm20\%$ for Athénan). For example, in Connect6, Athénan learns
55 times more states, makes 50 times less neural evaluations, and makes
3 times less state evaluations.
\label{data_stats}}
}{\footnotesize\par}
\end{table*}

\section{Final Discussion\label{sec:discusion}}

In \citep{cohen2020learning}, Athénan has been compared to other reinforcement
learning algorithms without knowledge that do not use learned policy.
In particular, using tree learning gives better results than root
learning or terminal learning (the AlphaZero / Polygames learning
technique for state values). As a reminder, tree learning learns the
entire partial search tree of the analyzed state while root learning
and terminal learning only learns the target value for the analyzed
state. More precisely, in the experiments of \citep{cohen2020learning},
the use of tree learning is always better than the use of terminal
learning and for almost all 9 tested games, tree learning improves
the final win rate by at least $40\%$. In this new article, we have
seen that tree learning can generate about 296
times more learning data than terminal learning (used by AlphaZero
/ Polygames). This is one of the reasons that allows Athénan
to obtain better results and in particular to learn much faster, especially
at the start of the learning process. Note that tree learning could not be applied
naturally to AlphaZero / Polygames (see Section \ref{data_comparison}).
Moreover, the use of tree learning can lower performance with some search algorithms (for various reasons ; see \citep{cohen2020learning} for details). This is not the case with Descent Minimax, in particular thanks to its exploration which is both very deep and in best first.

The superior performance of Athénan is not only due to the use of tree learning.
In \citep{cohen2020learning}, Descent Minimax, the central algorithm of Athénan, gives
better results than Unbounded Minimax, which is, itself, better than $\alpha\beta$ 
and Monte Carlo Tree Search (the search algorithm of AlphaZero / Polygames).
More precisely, using Descent Minimax with tree learning rather than MCTS
with tree learning increases the win rate by at least $40\%$ for
all 9 tested games. 

This previous study lacked a comparison with the state of the art,
which uses a policy (contrary to the studied techniques in \citep{cohen2020learning}). 
The question then was: do the results against MCTS without policy
generalize to the state of the art  (i.e. to MCTS with a learned policy)?
This article thus fills this gap and allows
one to conclude that with a reasonable hardware and an accessible time, 
Athénan gives much better results than Polygames (see
Section \ref{sec:compar_learning}). In addition, at least in some context,
Athénan with one GPU is even more efficient than Polygames with
100 GPUs (see Section \ref{subsec:tournament-comparison}).

\section{Conclusion\label{sec:conclusion}}

In \cite{cohen2020learning}, a new 
 framework for reinforcement learning without knowledge, called Athénan, has been proposed. In particular, in \cite{cohen2020learning}, Athénan has been compared to different standard search algorithms and learning techniques from the literature (which does not use a policy), and it has been shown that Athénan  obtains much better performance. However, Athénan has not been compared against the state of the art of reinforcement learning without knowledge, i.e. MCTS combined with a learned policy, the standard entire architecture being called AlphaZero.  This lack of comparison is all the more critical as the use of policy in the AlphaZero framework
  increases the level of play considerably. A comparison with the state of the art AlphaZero  is thus essential to know if Athénan is better  or if it is only a useful algorithm when a policy cannot be used.

Therefore, in this paper, we have made the first comparison between Athénan and Polygames, a re-implementation of AlphaZero. In particular, we have shown that Athénan has much
better performances than Polygames.

Recall that Athénan is a Minimax approach different in many points
from AlphaZero. Their basic differences are as follows. Athénan  does not use
a policy. It is based on Unbounded Minimax variants instead of MCTS.
It learns  as learning target the (partial) minimax value of states instead of the endgame
value. Finally, Athénan learns the values of all
the states of the search tree built during the match, while AlphaZero
only learns the values of the states of the match (i.e. AlphaZero
only learns the data of the states sequence of the match, which is
a small subset of states of the game search tree).

In our experiments, we have compared and revealed the cost of generating the learning
data. Athénan generates for the
same duration 296 times more learned states than Polygames.
This result is all the more striking since Athénan performs less than
half as many state evaluations than Polygames (because contrary to Polygames, Athénan is programmed in Python and Athénan does not performed matches in parallel).

In addition, we have compared the win rates of the two zero learning algorithms by evaluating them against MCTS with UCT on a large number of games. Athénan
obtains much better results than Polygames. In particular, Athénan
 is about 7 times faster than Polygames without reinforcement heuristic
and much more than $30$ times faster with reinforcement heuristic. Moreover,
we have performed another win rates comparison at 
Hex 13, in the context of a longer training that lasted 113 days, 
by evaluating them against Mohex 2.0, the best freely available Hex program.
Athénan has obtained again much better results than Polygames.

Finally, we have made a last comparison at Othello $8$, Othello $10$,
and Breakthrough, against top Polygames networks, having won two gold
medals and one silver medal at the 2020 TCGA tournaments. These Polygames
networks have been trained for a week with over 100 GPUs. It is again
 Athénan  that get the best results on each game, although
its training only lasted $5$ days and only required the use of half a GPU.

In conclusion, all these experiments show that for many games, 
reinforcement learning with Athénan is widely
more efficient than with Polygames, at least for accessible
learning times and reasonable resources use.

Note to conclude that Athénan faced Polygames during
the 2020 Computer Olympiad and beat it at Othello $8$, Othello $10$,
and Breakthrough. Athénan also beat other re-implementations
of AlphaZero during this competition (at Surakarta and Clobber). In
fact, $5$ gold medals were won by Athénan for the following
games: Othello $10$, Breakthrough, Surakarta, Amazons, and Clobber.
This was the first time that the same algorithm has won so many gold
medals in the same year.

Athénan has again competed in the 2021 Computer Olympiad.
This time, it won $11$ gold medals (Hex $11$, Hex $13$, Hex $19$,
Havannah $8$, Havannah $10$, Othello $8$, Surakarta, Amazons, Breakthrough,
Brazilian Draughts, Canadian Draughts; there was no competition at
Othello $10$ and Clobber). Athénan notably beat Polygames at games where
they met (Hex $13$, Hex $19$, Havannah $8$, Havannah $10$).

Athénan has again competed in the 2022 Computer Olympiad.
This time, it won $5$ gold medals (Surakarta, Breakthrough,
Canadian Draughts, Santorini, and Ataxx). Thus, Athénan is still the defending champion at $13$ games.

\balance

\begin{acks}
Thanks to Nicholas Sowels for proofreading. This work was granted access to the HPC resources of IDRIS under the allocation 2020-AD011011461, AD011011461R1, AD011011461R2, AD011011461R3, and 2020-AD\-011011714 made by GENCI. This work was supported in part by the French government under management of Agence Nationale de la Recherche as part of the “Investissements d’avenir” program, reference ANR19-P3IA-0001 (PRAIRIE 3IA Institute). 
\end{acks}

\bibliography{aamas2023}
\bibliographystyle{ACM-Reference-Format}

\end{document}